\documentclass[11pt,a4paper]{article}
\usepackage[hyperref]{emnlp2020}
\usepackage{times}
\usepackage{latexsym}
\usepackage{graphicx}
\usepackage{url}

\usepackage{ctable} 

\usepackage{microtype}

\aclfinalcopy 
\def\aclpaperid{3169} 

\title{Query-Key Normalization for Transformers}

\author{Alex Henry, Prudhvi Raj Dachapally, Shubham Pawar, Yuxuan Chen \\
 Cyndx Technologies \\
 \texttt{\{alex.henry,prudhvi.dachapally,shubham.pawar,ethan.chen\}@cyndx.com} \\}

\date{}

\begin{document}
\maketitle
\begin{abstract}
Low-resource language translation is a challenging but socially valuable NLP task. Building on recent work adapting the Transformer's normalization to this setting, we propose \textsc{QKNorm}, a normalization technique that modifies the attention mechanism to make the softmax function less prone to arbitrary saturation without sacrificing expressivity. Specifically, we apply $\ell_2$ normalization along the head dimension of each query and key matrix prior to multiplying them and then scale up by a learnable parameter instead of dividing by the square root of the embedding dimension. We show improvements averaging 0.928 BLEU over state-of-the-art bilingual benchmarks for 5 low-resource translation pairs from the TED Talks corpus and IWSLT'15.\footnote{Code to reproduce our experiments is available at \url{https://github.com/CyndxAI/QKNorm}}
\end{abstract}

\section{Introduction}

The Transformer \citep{vaswani2017attention} remains the architecture of choice for machine translation. Since its introduction, various architectural and functional modifications have been made to improve its performance on NMT datasets \citep{ahmed2017weighted,zhang2018accelerating,wang2019learning,dai2019transformer,zhao2019explicit}. Translating low-resource languages presents special challenges. Recent strategies for adapting Transformers to this socially valuable task include exploiting transfer learning with many-to-many multilingual models \citep{aharoni-etal-2019-massively}, reducing model depth \citep{van2020optimal}, and adding a regularization penalty for diverging from the predictions of a monolingual language model pretrained on the target language \citep{baziotis2020language}. This paper builds on recent work on layer normalization for low-resource language pairs, introducing a normalization technique that tries to keep the input to softmax attention within an appropriate range.

\paragraph{Layer normalization.} For Transformers and other NLP models, layer normalization \citep{ba2016layer} yields significantly better performance than batch normalization \citep{ioffe2015batch}, in part because NLP models tend to exhibit greater variance in batch statistics during training, for example compared to computer vision \citep{shen2020powernorm}. Layer normalization boosts performance in deeper networks chiefly by controlling their gradients \citep{xu2019understanding}. It re-scales and re-centers activation distributions (though re-centering may be unnecessary, see \citealt{NIPS2019_9403}). The type of normalization used and the placement of that normalization within the Transformer are both crucial to Transformer performance \citep{nguyen+salazar:iwslt2019}.

\paragraph{Softmax attention.} Given a matrix $X$ embedding a sequence of tokens, attention transforms each embedding into a mixture of itself and other elements of the sequence according to the importance of their connections for the modeling task at hand. In the case of multihead self-attention, the vectors of $X$ are projected linearly into Query, Key and Value matrices. The operation
\begin{equation}
\textnormal{softmax}(QK^{T})
\end{equation}
defines a distribution for each token over all the others in its sequence that sums to 1. Multiplying by $V$ then yields a new matrix where the embedding of each token is a weighted average of the vectors in $V$.

\citet{richter2020normalized} propose replacing the softmax function in attention because it constrains attention’s output to the convex hull spanned by the vectors in $V$, limiting model flexibility. For the softmax over the vocabulary in next word prediction, \citet{demeter-etal-2020-stolen} find that the norms of word embeddings drown out their angular displacements, with the consequence that words with smaller norms are systematically less likely to be predicted.

In this work, we replace the dot product inside of softmax attention with cosine similarity scaled up by a learnable parameter. This technique yields improved performance in low-resource bilingual translation, which we conjecture is because it binds $QK^{T}$ to a narrower range in a way that makes it easier to learn more diffuse attention patterns wherever these prove valuable.

\section{Background}

\citet{nguyen+salazar:iwslt2019} achieve state-of-the-art bilingual performance on 5 low-resource translation pairs from the TED Talks \citep{Ye2018WordEmbeddings} and IWSLT'15 \citep{Cettolo2015TheI2} corpora. This work builds directly on theirs, applying our technique to the same 5 benchmarks.  Their model combines three normalization techniques that we describe below: \textsc{FixNorm} \citep{nguyen-chiang-2018-improving}, \textsc{PreNorm} \citep{klein2017opennmt,domhan2018much,vaswani2018tensor2tensor,chen2018best}, and \textsc{ScaleNorm}, which they introduce as a replacement for layer normalization. They report that each technique contributes about 0.3 BLEU for an average improvement of 1.1 BLEU across the test sets for their 5 language pairs.

\textsc{FixNorm} sets word embeddings to unit length, which aids rare word translation \citep{nguyen-chiang-2018-improving}. \textsc{PreNorm} simply changes the location of layer normalization within the Transformer architecture, applying it to the input to each sublayer instead of after the residual connection. Moving layer normalization ahead of the residual connection enhances stability because the residual path is allowed to stay an identity map, instead of contributing terms to the gradient that could cause it to explode or vanish \citep{wang2019learning,nguyen+salazar:iwslt2019}. Interestingly, \citet{nguyen+salazar:iwslt2019} find \textsc{PreNorm} to be superior in low-resource but not high-resource translation settings.

Lastly, \textsc{ScaleNorm} replaces layer normalization with $\ell_2$ normalization along the embedding dimension, multiplied by a learnable scalar parameter initialized with $\frac{1}{\sqrt{d}}$ (where $d$ is the embedding dimension; the same term is used in scaled dot product attention \citep{vaswani2017attention}). 

In other words, \textsc{ScaleNorm} applies $\ell_2$ normalization along the embedding dimension of $Q$, $K$ \emph{and} $V$, and it does so \emph{before} the input to multihead attention gets split into heads. 

Building on their work, we combine \textsc{FixNorm}, \textsc{PreNorm}, and vanilla layer normalization (\textsc{LayerNorm}) with a new technique we call query-key normalization (\textsc{QKNorm}), surpassing their model’s performance on each of the same 5 translation pairs by an average of 0.928 test BLEU. 

\textsc{QKNorm} applies $\ell_2$ normalization to $Q$ and $K$ \emph{only}, and it does so along the \emph{head} dimension (which is the same dimension as the embedding dimension, but \emph{after} multihead attention has split its input into separate heads).  $Q$ and $K$ thus become $\hat Q$ and $\hat K$, where the \emph{i}th row vector of $\hat Q$ (the \emph{i}th embedding in the sequence) is given by:
\begin{equation}
\hat q_i = \frac{q_i}{||q_i||}\label{eqn:q_head_dim_norm}
\end{equation}
The effect is to make each element of $QK^{T}$ the cosine similarity of the corresponding pair of contextual token representations instead of their dot product. This is similar to \citet{luo2018cosine}, who propose replacing the dot product in fully-connected networks between layer weights and previous layer outputs with cosine similarity.

Like \textsc{ScaleNorm}, we also multiply by a learnable parameter that we initialize according to a rule of thumb we describe below.  Unlike \textsc{ScaleNorm}, \textsc{QKNorm} complements \textsc{LayerNorm} rather than replacing it.

\begin{figure*}[ht]
\centering
\includegraphics[width=\textwidth]{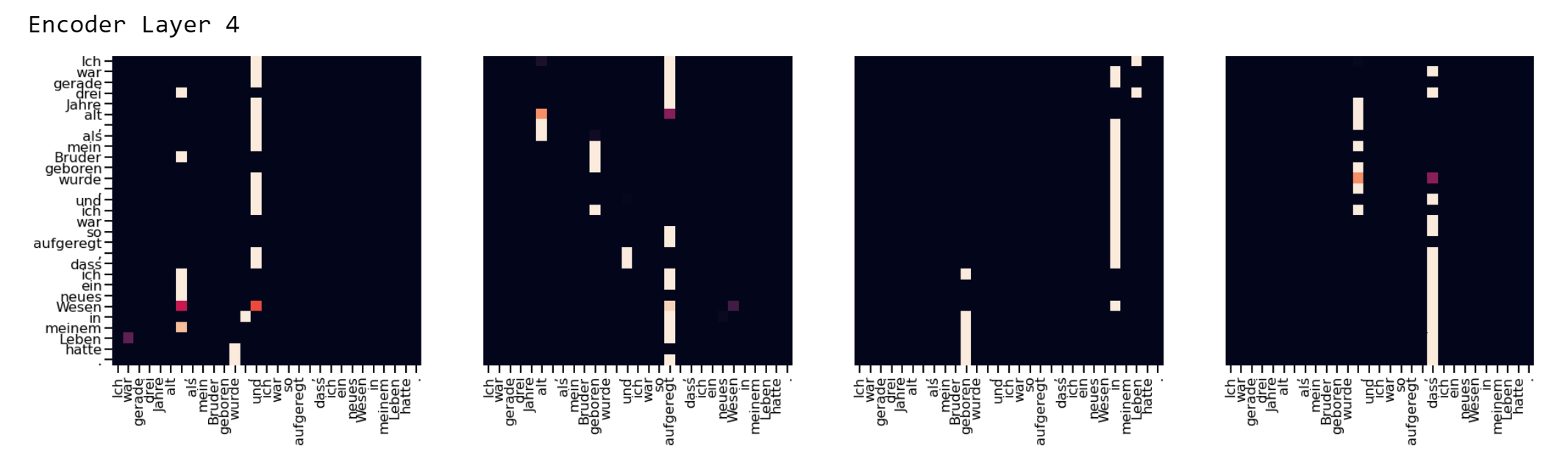}
\caption{\label{figure-1}
Scaled Dot Product Attention.  Self-attention heatmaps for 4 heads from one encoder layer displaying more ``concentrated'' attention, consistent with the conjecture that unnormalized dot products in $QK^{T}$ saturate the softmax and limit the attention patterns that can be learned.
}
\end{figure*}

\begin{figure*}[ht]
\centering
\includegraphics[width=\textwidth]{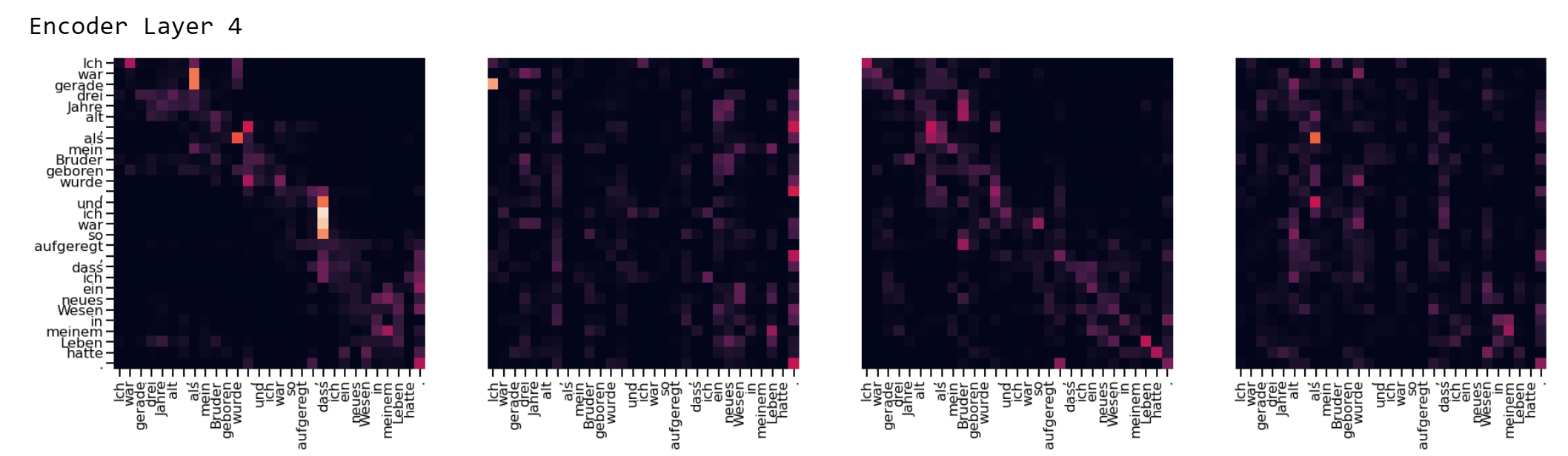}
\caption{\label{figure-2}
Query-Key Normalized Attention. Self-attention heatmaps of the same 4 heads in Figure \ref{figure-1}. \textsc{QKNorm} enables more diffuse attention patterns.
}
\end{figure*}

\begin{table*}
\centering
\resizebox{\textwidth}{!}{%
\begin{tabular}{c|c|c|c|c|c|c|c}
\specialrule{.1em}{.05em}{.05em} 
 &\textbf{Examples}&\textbf{Source + Target Tokens}&\textbf{Number of Parameters}&\textbf{Training Time (in hours)}&\textbf{Development BLEU}&\textbf{GPU}&\textbf{$L$}\\ \hline\hline
 \textbf{gl$\rightarrow$en}& 10k & 0.37M & 31,051,880 & 6 & 23.45 & T4 & 79 \\
 \textbf{sk$\rightarrow$en}& 61k & 2.32M & 48,356,907 & 11 & 31.34 & T4 & 75 \\
 \textbf{en$\rightarrow$vi}& 133k & 5.99M & 48,431,538 & 19 & 28.77 & T4 & 72 \\
 \textbf{en$\rightarrow$he}& 212k & 7.88M & 48,401,538 & 38 & 31.16 & T4 & 72 \\
 \textbf{ar$\rightarrow$en}& 214k & 8.09M & 48,499,512 & 26 & 37.94 & P100 & 75 \\ 
\specialrule{.1em}{.05em}{.05em} 
\end{tabular}
}
\caption{\label{tab:training_params}Summary of data and model training information. Number of examples and number of tokens taken directly from \citet{nguyen+salazar:iwslt2019}. $L$ is the 97.5th percentile sequence length across all training data sequences.}
\end{table*}

\begin{table*}
\centering
\scalebox{0.78}{
\begin{tabular}{c|c|c|c|c|c}
\specialrule{.1em}{.05em}{.05em} 
 &\textbf{en$\rightarrow$vi}&\textbf{ar$\rightarrow$en}&\textbf{en$\rightarrow$he}&\textbf{gl$\rightarrow$en}&\textbf{sk$\rightarrow$en}\\ \hline\hline
 \citet{nguyen+salazar:iwslt2019} & 32.79 & 36.09 & 28.28 & 22.01 & 32.58 \\
 \textsc{QKNorm}  + \textsc{LayerNorm} & \textbf{33.24} & \textbf{36.75} & \textbf{28.96} & \textbf{24.21} & \textbf{33.23} \\
\specialrule{.1em}{.05em}{.05em} 
\end{tabular}
}
\caption{\label{tab:bleu_comparison}Comparison of test BLEU \citep{papineni2002bleu}, scored using the \texttt{Moses} toolkit scripts provided in the repo for \citet{nguyen+salazar:iwslt2019}. $p < 0.01$ using bootstrap resampling \citep{koehn2004statistical}. Both architectures use \textsc{PreNorm}  and \textsc{FixNorm}. The \citet{nguyen+salazar:iwslt2019} architecture uses \textsc{ScaleNorm}  where we instead use vanilla layer normalization \citep{ba2016layer}, and scaled dot product attention where we use \textsc{QKNorm}.}
\end{table*}

\begin{table*}
\centering
\scalebox{0.78}{
\begin{tabular}{c|c|c|c|c|c}
\specialrule{.1em}{.05em}{.05em} 
 &\textbf{en$\rightarrow$vi}&\textbf{ar$\rightarrow$en}&\textbf{en$\rightarrow$he}&\textbf{gl$\rightarrow$en}&\textbf{sk$\rightarrow$en}\\ \hline\hline
 \citet{nguyen+salazar:iwslt2019} & 32.41 & 36.09 & 28.28 & 22.01 & 32.58 \\
 \textsc{QKNorm}  + \textsc{LayerNorm} & \textbf{33.18} & \textbf{36.75} & \textbf{28.96} & \textbf{24.21} & \textbf{33.22} \\
\specialrule{.1em}{.05em}{.05em} 
\end{tabular}
}
\caption{\label{tab:sacrebleu_comparison}Comparison of test BLEU \citep{papineni2002bleu}, scored using \textsc{SacreBLEU} \citep{post-2018-call}.}
\end{table*}

\section{Dot Products and the Softmax Function}

Softmax attends only to the differences between values. For example,
\begin{eqnarray}
&\textnormal{softmax}([760, 752, 750])& \nonumber \\
= &\textnormal{softmax}([12, 4, 2])&\nonumber\\
= &[0.99962, 0.00034, 0.00005].&\nonumber
\end{eqnarray}
Since the dot product is unbounded, differences between elements that may be insignificantly small on a relative basis can silence all other signals in the attention weights applied to $V$. We conjecture that this limits the complexity of the patterns that attention heads can learn.

The impact is more obvious in less sophisticated Transformer implementations (perhaps in part because subsequent advances have mitigated the same issue in different ways). Figures \ref{figure-1} and \ref{figure-2} show a heatmap comparison of encoder weights trained using the code for The Annotated Transformer\footnote{\url{https://nlp.seas.harvard.edu/2018/04/03/attention.html}}, the first with scaled dot product attention and the second with \textsc{QKNorm}.

The models containing these encoders were trained for 10 epochs on IWSLT 2016 \emph{de$\rightarrow$en} \citep{cettolo2016iwslt} using the Annotated Transformer implementation, with the baseline model scoring 19.4 BLEU and the \textsc{QKNorm} model scoring 24.33 BLEU on the test set, computed with the \texttt{SacreBLEU} Python package \citep{post-2018-call}. 

Though this heatmap comparison is obviously not systematic, we think the visual at least provides a plausible intuition for the incremental gain this technique achieves, with scaled dot product attention exhibiting the kind of ``winner-take-all'' behavior we would expect from a softmax near saturation.  

In comparison to dot products, cosine similarities are bounded by $[-1, 1]$ which creates the opposite problem as input to softmax -- the differences between values are too small for softmax to let the model effectively ignore connections between words it should not attend to. Instead of dividing by $\sqrt{d}$ as in scaled dot product attention we scale up using a learnable parameter that we initialize with a value that depends on the length of the sequences in the training data (and hence on the number of elements in $QK^{T}$): 
\begin{equation}
g_0 = \log_2(L^2-L)
\label{eqn:g_0}
\end{equation}
where $L$ is the 97.5th percentile sequence length across all training data sequences for source and target. 

The attention operation thus changes from
\begin{equation}
\textnormal{softmax}(\frac{QK^{T}}{\sqrt{d}})V
\label{eqn:scaled_attention}
\end{equation}
to
\begin{equation}
\textnormal{softmax}(g * \hat Q\hat K^{T})V
\label{eqn:qk_attention}
\end{equation}
where $\hat Q$ and $\hat K$ are $Q$ and $K$ with $\ell_2$-normalization applied along their head dimensions and $g$ is a learnable scalar parameter initialized with $g_0$ as computed in (\ref{eqn:g_0}).

\section{Experiments and Results}

We follow the implementation in the repository for \citet{nguyen+salazar:iwslt2019}, both in replicating their performance and as a starting point for our version (and also for computing BLEU as reported in Table~\ref{tab:bleu_comparison}).\footnote{\url{https://github.com/tnq177/Transformers_without_tears}} We train on the same 5 low-resource translation pairs as \citet{nguyen+salazar:iwslt2019}: 4 from the TED Talks corpus \citep{Ye2018WordEmbeddings}\footnote{\url{http://phontron.com/data/ted_talks.tar.gz}} -- Arabic, Slovak, and Galician translated to English, and English translated to Hebrew -- and 1 from the IWSLT'15 corpus \citep{Cettolo2015TheI2}, English to Vietnamese.  The repository for \citet{nguyen+salazar:iwslt2019} provides the tokenized text they used for English to Vietnamese.

\paragraph{Tokenization and BLEU.}Apart from BPE \citep{sennrich2015neural}, their repository does not include the code they used for tokenization, so for the other 4 language pairs we used the tokenization script from the repository for \citet{Ye2018WordEmbeddings}.\footnote{\url{https://github.com/neulab/word-embeddings-for-nmt/blob/master/ted_reader.py}}

The repository for \citet{nguyen+salazar:iwslt2019} includes two \texttt{Moses}\footnote{\url{https://github.com/moses-smt/mosesdecoder}} scripts for scoring BLEU, \texttt{multi-bleu.perl} and \texttt{multi-bleu-detok.perl}. We can't use \texttt{multi-bleu.perl} for the 4 TED Talks pairs without being able to replicate their tokenization because scores from that script are not comparable when there are differences in tokenization, unlike \texttt{multi-bleu-detok.perl} \citep{post-2018-call}. We use \texttt{multi-bleu.perl} to score \emph{en$\rightarrow$vi} (since we have their preprocessed text for this pair) and \texttt{multi-bleu-detok.perl} to score the 4 TED Talks pairs.

For additional confirmation, we also score all models using \texttt{SacreBLEU} \citep{post-2018-call} after detokenizing with NLTK's \texttt{TreebankWordDetokenizer} \citep{Loper02nltk:the}. These scores are reported in Table~\ref{tab:sacrebleu_comparison}.  All the detokenized BLEU scores from Table 2 are basically unchanged in Table 3, with the exception of  \emph{en$\rightarrow$vi}.  The best scores for the baseline model we could get on  \emph{en$\rightarrow$vi} were 32.48 for \texttt{Moses} \texttt{multi-bleu.perl} and 32.41 for \texttt{SacreBLEU}, though in Table 2 we report the \texttt{multi-bleu.perl}  score from \citet{nguyen+salazar:iwslt2019}, 32.79.   Our model's score for the same pair comes in 0.06 BLEU lower as well.

Following the \citet{nguyen+salazar:iwslt2019} repository, we perform BPE using \texttt{fastBPE}\footnote{\url{https://github.com/glample/fastBPE}}.  We also use the same \texttt{Moses} code for bootstrap resampling \citep{koehn2004statistical}.

\paragraph{Model hyperparameters.}Although \textsc{PreNorm} has been shown to make warmup less important for Transformers using scaled dot product attention \citep{nguyen+salazar:iwslt2019,xiong2020layer}, we obtained our best results using 8,000 steps of linear warmup.  How much linear warmup matters for \textsc{QKNorm} and why it matters are both subjects for further investigation.  We used the same validation-based decay scheme as \citet{nguyen+salazar:iwslt2019} and allowed models to train until they had reached the minimum learning rate.  For all other model hyperparameters and preprocessing settings we followed \citet{nguyen+salazar:iwslt2019} and the code in the lead author's GitHub repository.  As in their repository, we calculate test BLEU on the translation from the epoch with the highest validation BLEU.

\paragraph{Results.}Incorporating \textsc{QKNorm} and using layer normalization instead of \textsc{ScaleNorm}  boosted performance by an average of 0.928 BLEU across the test sets for the 5 translation pairs. On IWSLT'15 \emph{en$\rightarrow$vi}, our \texttt{SacreBLEU} test score of 33.18 is only 0.09 BLEU lower than \citet{provilkov-etal-2020-bpe}, who use BPE-dropout to increase BLEU 1.49 over the same model with vanilla BPE.

\section{Conclusion}
In this paper, we introduced a normalization technique that modifies the attention mechanism in Transformers and demonstrated its utility for low-resource bilingual translation by building it into an existing Transformer implementation with state-of-the-art performance on 5 low-resource language pairs.  \textsc{QKNorm} improves performance for each of the 5 pairs, with an average test BLEU increase of 0.928.  We pointed to possible explanations for its effectiveness but identifying exactly where it helps and why requires further research.  First, we plan to combine our approach with the fairseq Transformer implementation \citep{ott2019fairseq} and apply it to the \textsc{FLoRes} dataset \citep{guzman-etal-2019-flores}, investigating the effect of \textsc{QKNorm} on the optimal depth, number of attention heads, and warmup schedule for low-resource translation, in combination with recent advances like BPE-dropout \citep{provilkov-etal-2020-bpe}.  Next, we plan to look at high-resource settings to see whether the benefits of query-key normalization dissipate with access to more training data.  Lastly, we intend to study how \textsc{QKNorm} impacts what attention heads actually learn, adapting methods from BERT attention studies such as  \citet{clark-etal-2019-bert}.

\section*{Acknowledgments}

The authors would like to thank the reviewers for their valuable and insightful comments, and Toan Q. Nguyen for helpful clarifications and suggestions along the way.

\bibliographystyle{acl_natbib}
\bibliography{emnlp2020}

\appendix

\section*{Appendix}
\label{sec:appendix}

\section{Varying the Number of Heads}
In Table~\ref{number_of_heads}, we show the performance of \textsc{QKNorm} on the \emph{en$\rightarrow$vi} test set varying the number of heads. Even when the number of heads is 32 (with head dimension 16), the performance remains stable.

\begin{table}
\centering
\scalebox{0.9}{
\begin{tabular}{l|l}
\specialrule{.1em}{.05em}{.05em} 
\textbf{Number of Heads} & \textbf{Test BLEU} \\ \hline\hline
2 & 32.40 \\
4 & 33.16 \\
8 & 33.24 \\
16 & 32.42 \\
32 & 32.30 \\
\specialrule{.1em}{.05em}{.05em} 
\end{tabular}
}
\caption{\label{number_of_heads} IWSLT'15 \emph{en$\rightarrow$vi} test BLEU for \textsc{QKNorm} varying the number of attention heads.}
\end{table}

\section{Equation \ref{eqn:g_0}}
Intuitively, longer sequences require more scaling to make it at least possible for the maximum values in $QK^{T}$ to softmax to 1.  We arrived at Equation \ref{eqn:g_0} empirically by applying softmax to similarity matrices of word vectors scaled up with various heuristics.  Like ${\sqrt{d}}$ in scaled dot product attention \citep{vaswani2017attention}, Equation \ref{eqn:g_0} is a rule of thumb but it initializes a learnable parameter.  

We determined the best value of $L$ in Equation \ref{eqn:g_0} by running the \emph{en$\rightarrow$vi} translation task with different percentile values. Table~\ref{percentile} shows the results from those experiments.

\begin{table}
\centering
\scalebox{0.9}{
\begin{tabular}{l|l}
\specialrule{.1em}{.05em}{.05em} 
\textbf{Percentile} & \textbf{Test BLEU} \\ \hline\hline
75th & 32.58 \\
90th & 32.89 \\
92.5th & 32.64 \\
95th & 33.13 \\
97.5th & 33.24 \\
99th & 32.64 \\
Maximum Word Count & 33.10 \\
\specialrule{.1em}{.05em}{.05em} 
\end{tabular}
}
\caption{\label{percentile} IWSLT'15 \emph{en$\rightarrow$vi} test BLEU for \textsc{QKNorm} varying the training set word count percentile used to initialize the learnable scaling factor $g$.}
\end{table}

\section{Ablation Experiments}

Table~\ref{ablation_experiments} shares test performance on \emph{en$\rightarrow$vi} when we ablate specific components of \textsc{QKNorm}.  The biggest performance drop in these experiments comes from omitting $g$, the learnable scaling factor.  This is unsurprising because if we don't scale up $\hat Q\hat K^{T}$ its values are all within $[-1, 1]$ and softmax is a function of the differences between values.  

\begin{table}
\centering
\scalebox{0.9}{
\begin{tabular}{l|l}
\specialrule{.1em}{.05em}{.05em}
\textbf{Experiment} & \textbf{Test BLEU} \\ \hline\hline
Without $g$ & 24.53 \\
Without \textsc{LayerNorm} & 31.56 \\
Without \textsc{FixNorm} & 32.63 \\
Without \textsc{FixNorm} or \textsc{PreNorm} & 32.20 \\
$\ell_2$-normalizing $V$ along with $Q$ and $K$ & 32.34 \\
\specialrule{.1em}{.05em}{.05em} 
\end{tabular}
}
\caption{\label{ablation_experiments} Ablation Experiments.}
\end{table}

\end{document}